\documentclass[runningheads]{llncs}
\usepackage[T1]{fontenc}
\usepackage{graphicx,verbatim}
\usepackage{multirow}
\usepackage{tabularx}
\usepackage{booktabs}
\usepackage{caption} 
\usepackage{tikz}
\usepackage{url}
\usepackage{adjustbox}
\usepackage{amsmath}
\usepackage{amssymb}
\begin{document}
\title{Flow Matching for Medical Image Synthesis: Bridging the Gap Between Speed and Quality}
%

\author{Milad Yazdani\inst{1}\and 
Yasamin Medghalchi\inst{1}\and  
Pooria Ashrafian\inst{1}\and 
Ilker Hacihaliloglu\inst{1}\and 
Dena Shahriari\inst{1}
}  

\authorrunning{M. Yazdani et al.}
\institute{University of British Columbia, Vancouver, Canada\\  
\email{milad.yazdani@ece.ubc.ca}
}

\maketitle              
\begin{abstract}
Deep learning models have emerged as a powerful tool for various medical applications. However, their success depends on large, high-quality datasets that are challenging to obtain due to privacy concerns and costly annotation. Generative models, such as diffusion models, offer a potential solution by synthesizing medical images, but their practical adoption is hindered by long inference times. In this paper, we propose the use of an optimal transport flow matching approach to accelerate image generation. By introducing a straighter mapping between the source and target distribution, our method significantly reduces inference time while preserving and further enhancing the quality of the outputs. Furthermore, this approach is highly adaptable, supporting various medical imaging modalities, conditioning mechanisms (such as class labels and masks), and different spatial dimensions, including 2D and 3D. Beyond image generation, it can also be applied to related tasks such as image enhancement. Our results demonstrate the efficiency and versatility of this framework, making it a promising advancement for medical imaging applications.
Code with checkpoints and a synthetic dataset (beneficial for classification and segmentation) is now available on: \url{https://github.com/milad1378yz/MOTFM}.


\end{abstract}
\section{Introduction}
Over the past decade, artificial intelligence (AI), especially deep learning (DL), has significantly advanced disease detection and segmentation from medical images \cite{ting2018ai}. However, building reliable AI models for medical image analysis requires large, diverse datasets, which are hard to obtain due to privacy restrictions, rare diseases, and inconsistent diagnostic labels \cite{singh2020current}.
One solution is to generate synthetic data to augment existing datasets \cite{frid2018gan}. Deep generative models have shown promising results in various medical applications, enabling more robust training for machine learning models.
Early generative models like Generative Adversarial Networks (GANs) \cite{goodfellow2014generative} have been widely used for medical image synthesis across modalities, including echocardiographic imaging \cite{tiago2022data}. Conditional GANs incorporate additional inputs for greater control, as in SPADE \cite{park2019semantic}, which uses semantic layouts and has been applied to CT liver volumes, retinal fundus images, and cardiac cine-MRI \cite{skandarani2023gans}. Despite generating high-quality images, GANs often lack diversity and suffer from training instability and mode collapse without careful tuning \cite{nichol2021improved}. To address these limitations, diffusion models, particularly Denoising Diffusion Probabilistic Models (DDPM) \cite{ho2020denoising}, have emerged as a powerful alternative. By formulating the image synthesis process as a continuous-time diffusion governed by stochastic differential equations (SDEs), DDPMs address some limitations of GANs, achieving superior performance in both image quality and diversity \cite{dhariwal2021diffusion}. Recent studies show that these diffusion-based approaches even outperform GANs in generating medical images in various applications
such as echo \cite{ashrafian2024vision}.
Diffusion models incorporate various conditioning mechanisms for enhanced control. Prompt-based conditioning, like Latent Diffusion Models (LDM) \cite{Rombach_2022_CVPR}, has been applied to Breast MRI and head CT synthesis \cite{10497435}, while mask- or image-based conditioning, as in ControlNet \cite{zhang2023adding}, has been used for colon polyp synthesis, showcasing their flexibility in medical imaging.
Building on these advancements, diffusion models have also been optimized for efficiency. Deterministic variants like DDIM \cite{song2020denoising} reframe the stochastic process as an ordinary differential equation (ODE), significantly reducing inference steps while maintaining performance. However, despite their advantages, diffusion models still rely on iterative denoising through numerical ODE/SDE solvers, resulting in slow inference times. While DDIM mitigates this by reducing the number of steps, the underlying iterative solvers remain computationally demanding.  In parallel, an alternative family of generative models, known as Flow Matching, emerged. Unlike diffusion models (which maximize a variational lower bound), flow matching directly maximizes the likelihood, potentially improving generalization\cite{lipman2022flow}. Additionally, flow matching methods are more flexible, allowing for diverse path definitions such as Gaussian, affine, and linear trajectories \cite{lipman2022flow}.
A key advancement in this family is flow matching with optimal transport \cite{liu2022flow}, which provides a direct mapping between the source (typically noise) and the target distribution. Unlike diffusion models, which rely on iterative sampling, flow matching with optimal transport enables significantly faster sampling by constructing an efficient transport plan. 
Notably, optimal transport flow matching has shown remarkable performance in natural image generation \cite{lipman2022flow} and enhancement \cite{zhu2024flowie}, yet its potential for medical imaging remains largely unexplored. To the best of our knowledge, this is the first work leveraging flow matching with optimal transport for medical image synthesis.
\textbf{Our key contributions are as follows:}
\textbf{1)} We present the first medical image synthesis framework leveraging \textit{Optimal Transport Flow Matching}, significantly accelerating inference while outperforming diffusion-based models in image quality.
\textbf{2)} We evaluate its effectiveness across Unconditional, Class-Conditional, and Mask-Conditional Image Generation, demonstrating its 
robustness, and versatility across diverse generative tasks.
\textbf{3)} Our method adapts to different medical imaging modalities (e.g., ultrasound, MRI) and spatial dimensions (2D, 3D), ensuring broad applicability. 
\textbf{4)} The proposed approach supports end-to-end training, eliminating the need for complex post-processing steps and simplifying the learning pipeline.
\noindent This study opens a new pathway for medical image generation, demonstrating that flow matching with optimal transport is an effective and efficient alternative to traditional generative models in the medical field.

\section{Method}
\subsection{Preliminaries} 
\begin{figure}
    \centering
    \includegraphics[width=1.0\textwidth]{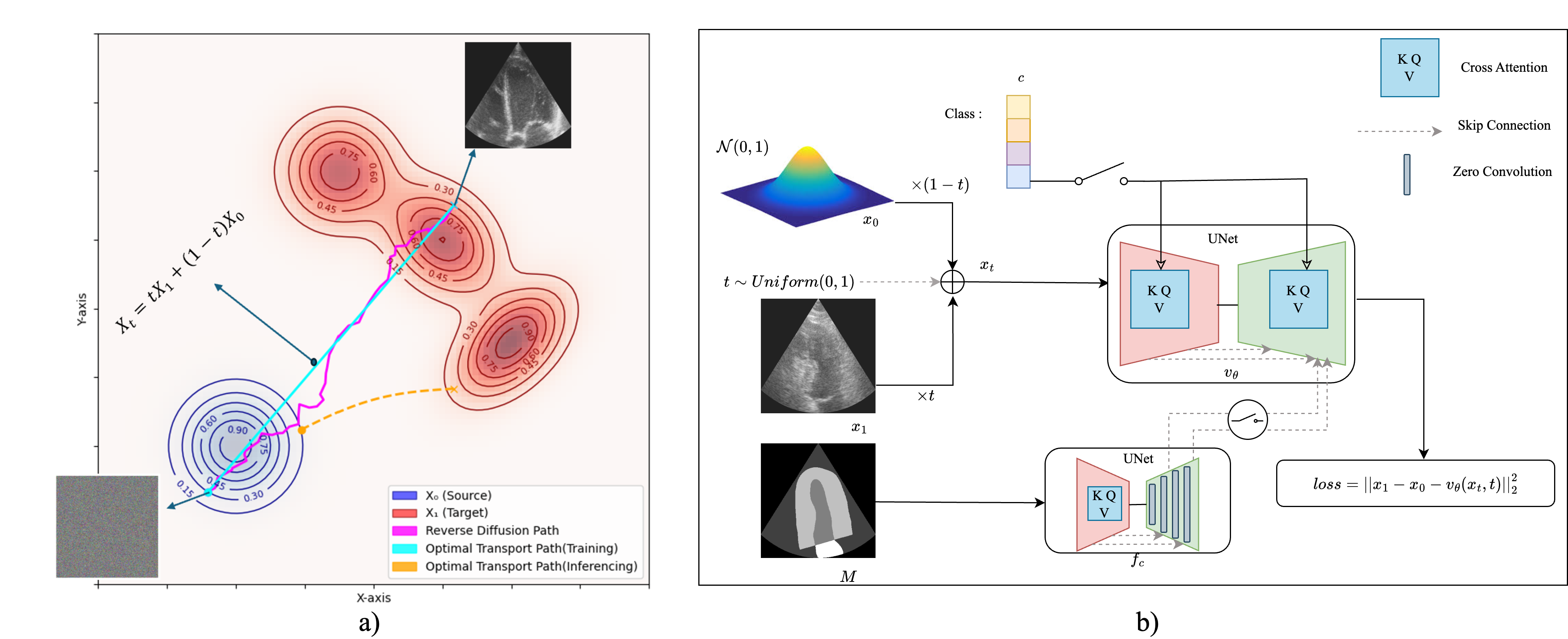}
    \caption{a) The figure illustrates transitions from source \(x_0\) (blue) to target \(x_1\) (red). Diffusion models map noisy samples to targets (magenta), while flow matching provides a more efficient path (cyan for training, dashed orange for inference). The contour map represents probability density.
    b) MOTFM framework with different conditioning strategies.}
    \label{rf_diff}
\end{figure}

Both flow matching with optimal transport and diffusion models \cite{ho2020denoising} aim to generate data from a complex target distribution \( X_1 \) starting from a simple Gaussian prior \( X_0 \). Diffusion models achieve this by modeling the transformation as a continuous-time stochastic differential equation (SDE), where a neural network estimates the drift term. These SDEs can be reformulated as probability flow ordinary differential equations (ODEs) \cite{song2020denoising}, preserving marginal distributions while enabling faster inference. In diffusion models, data gradually transitions from \( X_1 \) to \( X_0 \) by introducing Gaussian noise, \(\epsilon\), at each step, $x_t = \sqrt{\bar{\alpha}_t}\, x_1 + \sqrt{1 - \bar{\alpha}_t}\, \epsilon.$ such that at the final time step \( T \) the data distribution becomes pure noise, \(x_0\), and the model learns to estimate \(\epsilon\). The training objective minimizes the noise prediction error, $\mathcal{L}_{\text{diff}} = \mathbb{E}_{\epsilon \sim \mathcal{N}(0,\,I)} \left[ \|\epsilon - \epsilon_\theta(x_t,t)\|_2^2 \right].$
Note that in practice the naming in diffusion models is often reversed (e.g., \(x_{T}\) is noise and \(x_0\) is the target); for consistency, we denote \(x_0\) as the noise (source) and \(x_1\) as the target, as illustrated in Fig.~\ref{rf_diff}.
Since the diffusion process follows a highly non-linear trajectory (Fig.~\ref{rf_diff}.a), inference requires multiple iterative steps. In contrast, flow matching with optimal transport addresses this inefficiency by defining a straight-line (or nearly straight) transformation between \( X_0 \) and \( X_1 \) that approximates the optimal transport map under a quadratic cost. Specifically, it models data as $x_t = t\, x_1 + (1-t)\, x_0$.
and trains a neural network to estimate the velocity field \( v_\theta(x_t,t) \) such that ideally $v_\theta(x_t,t) = x_1 - x_0.$
The corresponding loss function is $\mathcal{L}_{\text{OTFM}} = \mathbb{E}_{x_0, x_1} \left[ \| (x_1 - x_0) - v_\theta(x_t,t) \|_2^2 \right].$
From the perspective of optimal transport, this formulation seeks to minimize the transport cost between the source and target distributions by matching the learned velocity with the true optimal transport velocity.

\noindent During inference, samples are generated by solving the differential equation, $\frac{dx_t}{dt} = v_{\theta}(x_t,t).$ 
This direct mapping allows the model to theoretically recover \( X_1 \) from \( X_0 \) in a single step. In practice, however, the inferencing path is not strictly linear, and a minimal number of steps is still required far fewer than in diffusion models \cite{liu2022flow}. Mathematically, by leveraging the structure of optimal transport, flow matching minimizes the discrepancy between the learned and optimal velocity fields, effectively bridging the two distributions with near-optimal efficiency.
\subsection{Medical Optimal Transport Flow Matching (MOTFM)}
Our framework, Medical Optimal Transport Flow Matching (MOTFM), generates synthetic images using a UNet backbone \cite{ronneberger2015u} with attention layers \cite{vaswani2017attention} to estimate velocity in flow matching. To improve memory efficiency, we employ flash attention \cite{dao2022flashattention} instead of standard attention.
Flash attention is a GPU-optimized technique that reduces memory usage by loading queries, keys, and values only once and performing block-wise computations, thereby accelerating training and inference.
As shown in Fig.~\ref{rf_diff}.b, training involves progressively adding Gaussian noise to images with optimal transport. The UNet learns to estimate the velocity field that maps noisy inputs to the original image. The loss between the predicted and actual velocity fields is used to optimize network parameters. This approach is generalizable to both unconditional and conditional image generation, as described below.\\
\noindent\textbf{a) Unconditional Image Generation}
As shown in Fig.~\ref{rf_diff}.b, when no conditioning is provided, all switches are off, generating an image unconditionally.
\noindent\textbf{b) Class-conditional Image Generation} The generation process can be guided by incorporating a one-hot class vector, indicating class number, into the UNet via the cross-attention mechanism, enabling generating images with a specific class.
\noindent\textbf{c) Mask-Conditional Image Generation}
Our method can optionally be guided by masks to generate images, which is particularly important if the downstream task is image segmentation.
We introduce an additional UNet, $f_c$ (Fig.~\ref{rf_diff}.b), to encode the mask. The decoder of $f_c$ employs zero convolutions, and skip connections are established between the decoder of $f_c$ and $v_\theta$ to incorporate mask-based guidance. Notably, during training, both UNets are updated end-to-end in contrast with ~\cite{zhang2023adding}. Furthermore, this mask-conditioning approach can be seamlessly combined with class-conditioning. This pipeline can be extended to other image-to-image translation tasks. 
\section{Results}
\noindent\textbf{Experiment Settings.}
All generative models were trained with the Adam optimizer with learning rate 1e-4 for 200 epochs on an NVIDIA RTX 4090. For classification and segmentation, the same optimizer was used for 50 epochs, selecting the best model based on validation performance.
 
\begin{figure}[!h]
\centering
\includegraphics[width=0.7\textwidth]{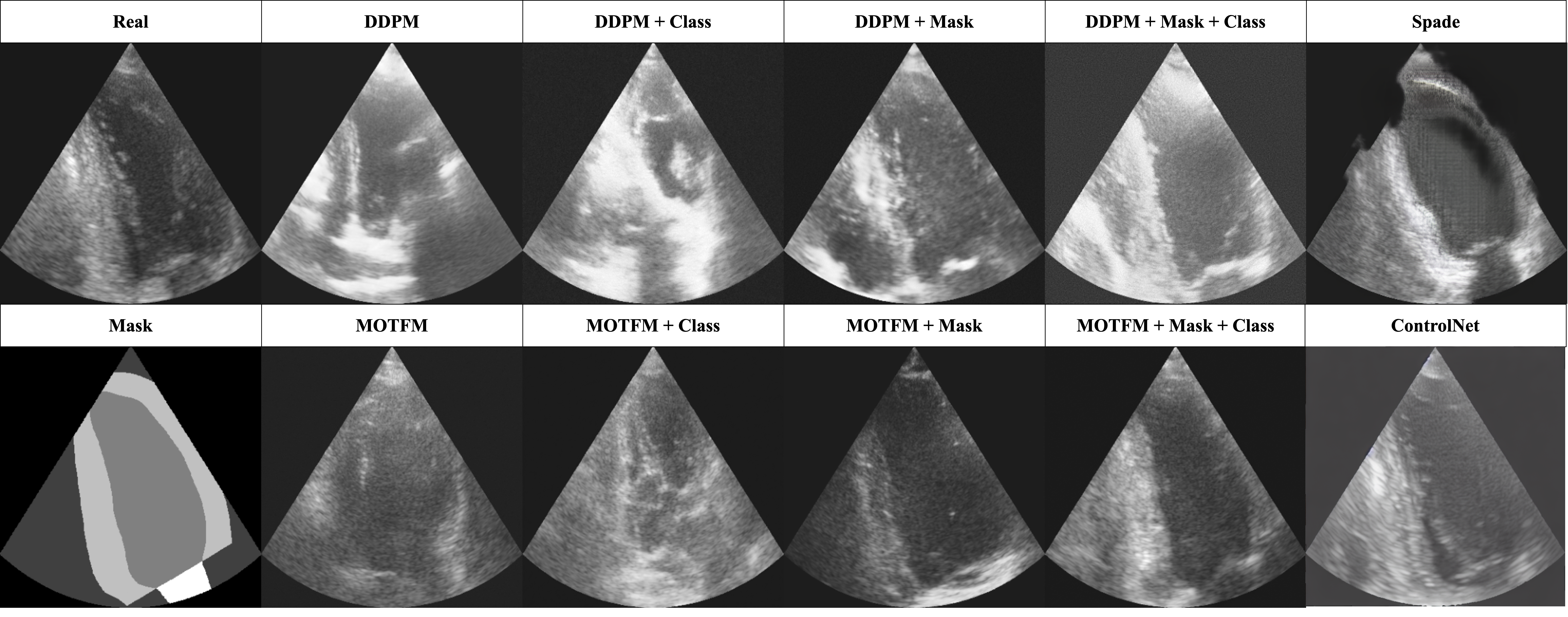 } 
\includegraphics[width=0.7\textwidth]{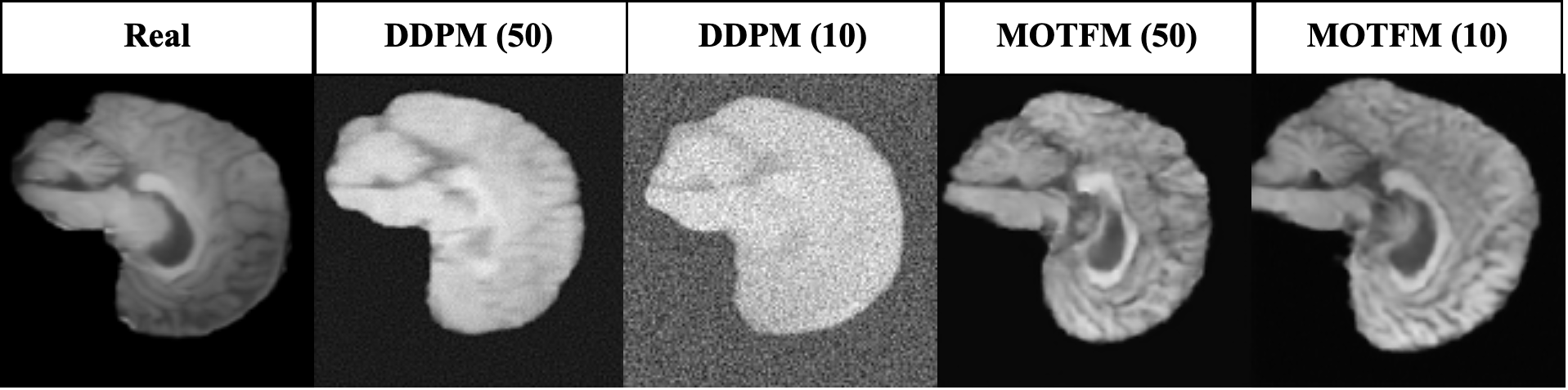} 
    \caption{Comparison of echocardiographic and brain MRI synthesis using DDPM and MOTFM, with SPADE and ControlNet applied only to echocardiography. The first two rows show echocardiographic images, while the last row presents brain MRI synthesis, with numbers in parentheses indicating inference steps.}
    \label{Examples}
\end{figure}

\noindent For 3D generation tasks, such as MRI synthesis, the overall approach remains consistent, with the primary modifications being the use of 3D layers in place of their 2D counterparts. \\
\noindent\textbf{Datasets.} 
We utilized two datasets, one of which is the \textbf{CAMUS echocardiography (echo) dataset} \cite{leclerc2019deep}, which contains 2D apical views of both two-chamber (2CH) and four-chamber (4CH) perspectives from 450 patients, covering both end-diastole (ED) and end-systole (ES) phases. Initially, a subset of images from 50 patients was randomly selected for the test split, while the remaining 400 patients were designated for training. Following the baseline approach \cite{stojanovski2023echo}, we further partitioned the dataset by assigning the first 50 patients to the validation set, leaving the remaining 350 patients for training. This resulted in a total of 1400 images for training and 200 images for validation. As a second dataset, we evaluate our pipeline on a different modality and dimension using the \textbf{3D MSD MRI Brain Dataset} \cite{antonelli2022medical}. The brain tumor dataset from the Medical Segmentation Decathlon (MSD) challenge \cite{antonelli2022medical} comprises 750 multiparametric MRI scans from patients diagnosed with glioblastoma or lower-grade glioma. Each scan includes T1-weighted (T1), post-Gadolinium contrast T1 (T1-Gd), T2-weighted (T2), and T2-FLAIR sequences. These images, collected from 19 institutions, were acquired using diverse clinical protocols and imaging equipment. For simplicity, we focus only on T1-weighted images in this study.\\
\noindent\textbf{Experiments.}  
We evaluate our pipeline on two mentioned datasets, comparing it to baselines, including our framework with DDPM (under various conditioning settings), SPADE, and ControlNet as a mask-guided approach. For efficiency, we used the DDIM scheduler \cite{nichol2021improved} during DDPM sampling. The echocardiography dataset is used for all conditioning methods, while the MRI dataset is used for unconditional generation.\\
\noindent\textbf{Qualitative Visualization.}
Fig.~\ref{Examples} presents image generation examples for echocardiography (first two rows) and MRI (last row) using different conditioning strategies for DDPM and MOTFM. In the echocardiography dataset, the mask is used solely for conditioning in the mask-based generation approach.  
Across both modalities, DDPM introduces a noticeable increase in brightness, deviating from the real data distribution, whereas MOTFM maintains a pixel distribution more consistent with real images. A similar brightness shift is also observed in the ControlNet-generated echocardiography images. To quantify this effect, Fig.~\ref{hist} shows the kernel density estimation (KDE) of pixel intensities for real, DDPM-generated, and MOTFM-generated echocardiographic images. The KDE plot highlights a higher density in the bright intensity range [150, 250] for DDPM, indicating a shift toward brighter outputs. This brightness shift, also observed in natural image generation \cite{corvi2023detection}, highlights the limitations of diffusion models in maintaining realistic intensity distributions. 
Furthermore, MOTFM outperforms DDPM, ControlNet, and SPADE in echocardiographic image quality. In MRI synthesis, 50-step MOTFM produces higher-quality images than DDPM, and even its 10-step output outperforms 50-step DDPM, highlighting its efficiency.
 \\

\noindent\textbf{Generation Evaluations (CAMUS Dataset).} To evaluate our framework, we compare MOTFM with baselines (mentioned in "\textbf{Experiments}") in echo image generation. Performance is evaluated using FID, CMMD, and KID (distribution distance), IS (sample diversity), and SSIM (structural similarity) \cite{FID,inception_score,jayasumana2024rethinking,xu2018empirical,wang2004image}.

Table~\ref{tab:echo_gen_metrics} presents the results for the CAMUS dataset, computed over 2000 generated images. MOTFM consistently outperforms baselines across most metrics.  Notably, one-step MOTFM outperforms 10-step DDPM and achieves the same order of performance as 50-step DDPM, demonstrating superior efficiency without compromising quality. While its slightly lower Inception Score suggests a trade-off between realism and diversity, MOTFM’s substantial gains in FID, SSIM, and CMMD establish it as a more effective and efficient alternative for echocardiographic image synthesis. We also compared inference times: 10-step MOTFM finishes in 1.47s, much faster than 50-step DDPM (6.20s) and approaching SPADE’s 0.68s, further underscoring its efficiency.\\

\noindent\textbf{Generation Evaluations (MSD MRI Brain Datase).}
To evaluate 3D image generation on the 3D MSD brain dataset, we compare MOTFM with DDPM for unconditional synthesis using 3D-FID, MMD (distribution distance), and MS-SSIM (structural similarity), following prior work on 3D brain MRI generation \cite{sun2022hierarchical,gretton2012kernel,wang2003multiscale,dorjsembe2024conditional}.
  
Table~\ref{tab:MRI_gen_metrics} reports results over 2000 generated samples, showing that MOTFM consistently outperforms DDPM. Notably, one-step MOTFM surpasses 50-step DDPM in MS-SSIM and MMD and achieves competitive results in the 3D-FID, demonstrating its efficiency and adaptability for 3D medical image synthesis.

\begin{table}[!h]
    \caption{Evaluation of Echocardiography Image Generation. D, M, and SD-M represent DDPM, MOTFM, and ControlNet, respectively. "-C", "-M", and "-CM" denote class, mask, and class + mask conditionings. The columns 1, 10, and 50 indicate the number of inference steps. The best results for each conditioning are highlighted in \textbf{bold}.}
    \label{tab:echo_gen_metrics}
    \centering
    \renewcommand{\arraystretch}{1.2}
    \begin{tabular}{|l|c|c|c|c|c|c|c|c|c|c|c|c|c|c|c|}
        \hline
        \scriptsize
         & \multicolumn{3}{c|}{\textbf{FID $\downarrow$}} & \multicolumn{3}{c|}{\textbf{SSIM $\uparrow$}} & \multicolumn{3}{c|}{\textbf{KID $\downarrow$}} & \multicolumn{3}{c|}{\textbf{CMMD $\downarrow$}} & \multicolumn{3}{c|}{\textbf{IS $\uparrow$}} \\
        \hline
        & 1 & 10 & 50 & 1 & 10 & 50 & 1 & 10 & 50 & 1 & 10 & 50 & 1 & 10 & 50 \\
        \hline
        \scriptsize{D} &1.9e2 
        & 22.83 & 1.84 
        & .00 & .08 & 0.29 
        &1.6e3 
        & 62.2 & 1.29 
        & 5.38 & 1.30 & 1.52 
        & 1.07 & \textbf{3.16} & 2.07 \\
        \scriptsize{M} & 3.04 & .16 & \textbf{.04} 
        & \textbf{.70} & .65 & .62 
        & 4.61 & 0.19 & \textbf{.03} 
        & 1.95 & .96 & \textbf{.50} 
        & 1.16 & 1.42 & 1.39 \\
        \hline
        \scriptsize{D-C}  &1.9e2 
        & 15.21 & 4.01 
        & .00 & .10 & 0.27 
        & 1.6e3 
        & 22.1 & 5.36 
        & 5.38 & 1.61 & 1.59
        & 1.07 & \textbf{2.84} & 1.92 \\
        \scriptsize{M-C}  & 1.93 & .08 & \textbf{.06} 
        & .62 & .64 &\textbf{0.65} 
        & 2.93 & \textbf{.07} & .08 
        & 2.15 & 1.73 & \textbf{.76}
        & 1.34 & 1.35 & 1.34 \\
        \hline
        \scriptsize{Spade} & \multicolumn{3}{c|}{.46} 
        & \multicolumn{3}{c|}{.54} 
        & \multicolumn{3}{c|}{.73} 
        & \multicolumn{3}{c|}{.46} 
        & \multicolumn{3}{c|}{1.73} \\
        \hline
        \scriptsize{D-M} &1.9e2 
        & 14.58 & 1.61 
        & .00 & .14 & .35 
        & 1.6e3
        & 24.1 & .75 
        & 5.39 & 1.62 & 1.57 
        & 1.07 & \textbf{2.36} & 1.72 \\
        \scriptsize{SD-M} & 5.67 & 2.25 & 1.82 
        & .57 & .56 & .63 
        & 8.87 & 2.37 & 1.99 
        & 3.42 & 0.42 & .39 
        & 1.70 & 1.76 & 1.67 \\
        \scriptsize{M-M} & 3.91 & .58 & \textbf{.22} 
        & \textbf{.72} & .67 & .66 
        & 5.81 & .75 & \textbf{.23} 
        & 1.29 & .16 & \textbf{.12} 
        & 1.30 & 1.28 & 1.23 \\
        \hline
        \scriptsize{D-CM} &1.9e2 
        & 25.66 & 7.98 
        & .00 & .15 & .28 
        &1.6e3 
        & 83.8 & 17.05 
        & 5.39 & 1.37 & 1.23
        & 1.07 & \textbf{2.67} & 2.13 \\
        \scriptsize{M-CM} & 3.21 & \textbf{.07} & \textbf{.07} 
        & .64 & .69 &\textbf{ .70} 
        & 5.01 & .05 & \textbf{.03}
        & 1.94 & .72 & \textbf{.64} 
        & 1.34 & 1.34 & 1.33 \\
        \hline
    \end{tabular}
\end{table}

\begin{table}[!ht]
    \centering
    \begin{minipage}{0.7\linewidth} 
        \centering
        \caption{Evaluation of Brain MRI Unconditional Image Generation. D and  M represent DDPM and MOTFM respectively. The columns 1, 10, and 50 indicate the number of inference steps. MMD values in the tables are divided by 1000 for readability. The best results are highlighted in \textbf{bold}.}
        \label{tab:MRI_gen_metrics}
        \renewcommand{\arraystretch}{1.2}
        \begin{tabular}{|l|c|c|c|c|c|c|c|c|c|}
            \hline
             & \multicolumn{3}{c|}{\textbf{3D-FID $\downarrow$}} & \multicolumn{3}{c|}{\textbf{MS-SSIM $ \uparrow $}} & \multicolumn{3}{c|}{\textbf{MMD $ \downarrow $}}  \\
            \hline
            & 1 & 10 & 50 & 1 & 10 & 50 & 1 & 10 & 50 \\
            \hline
            \scriptsize{D} &146.47 & 51.68& 29.67&  .06&  .51&  .59
            & 39.8& 26.1& 4.28\\
            \hline
            \scriptsize{M} &32.10&9.27&\textbf{7.93}&.66&.77&\textbf{.77}
            &.51&.25&\textbf{.22}\\
            \hline
        \end{tabular}
    \end{minipage}%
    \hfill
    \hfill
    \begin{minipage}{0.30\linewidth} 
\includegraphics[width=\linewidth]{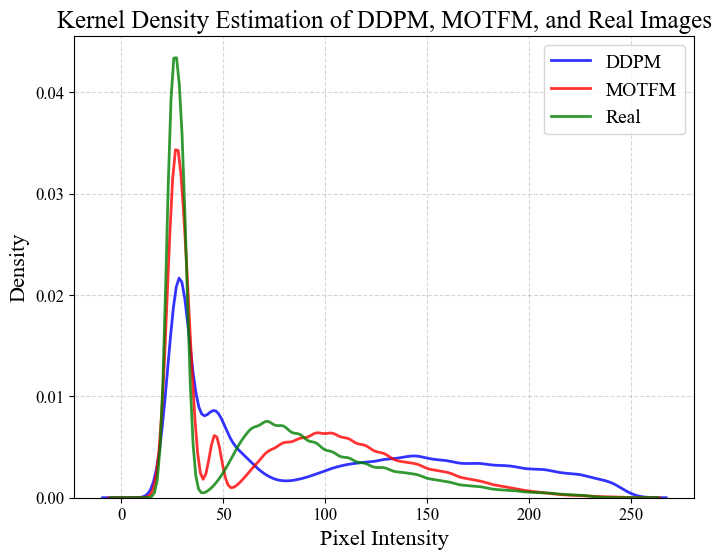} 
            \captionof{figure}{KDE Plot of Pixel Intensity Distributions for Generated and Real Echo Images.}
            \label{hist} 
    \end{minipage}
\end{table}

\noindent\textbf{Downstream Tasks.} To further evaluate the realism of our generated images, we assess their performance in downstream tasks, specifically classification and segmentation. We first train models (Table~\ref{tab:classification_metrics}) on real training images and evaluate them on the real test set of the CAMUS dataset. For segmentation, UNet-ResX refer to UNet models that use ResNetX as their encoder backbones. To compare, we train classifiers on only class-conditioned synthetic images and segmentors on only mask-conditioned synthetic images, ensuring dataset sizes match the real training set. All downstream models trained on synthetic data were also tested exclusively on real ultrasound data.  MOTFM generates images in just 10 steps, whereas DDPM requires 50 steps to achieve comparable quality. As shown in Table~\ref{tab:classification_metrics}, our 10-step MOTFM method outperforms the 50-step DDPM, yielding results closer to the original distribution. This demonstrates MOTFM’s improved fidelity and efficiency, making it a more effective approach for medical image synthesis.

\begin{table}[!h]
    \centering
    \caption{Classification and Segmentation Metrics for Echo}
    \label{tab:classification_metrics}
    \begin{tabular}{l l c c | l c c c c}
        \toprule
        \multicolumn{4}{c|}{\textbf{Classification}} & \multicolumn{5}{c}{\textbf{Segmentation}} \\
        \cmidrule(lr){1-4} \cmidrule(lr){5-9}
        \textbf{Model} & \textbf{Data} & \textbf{ACC $\uparrow$} & \textbf{F1 $\uparrow$} & \textbf{Model} & \textbf{Dice $\uparrow$} & \textbf{IOU $\uparrow$} & \textbf{HD $\downarrow$} & \textbf{ASD $\downarrow$} \\
        \midrule
        \multirow{3}{*}{ResNet18} 
            & Real       & 0.89  & 0.89  & \multirow{3}{*}{UNet-Res18} & 0.92  & 0.86   & 16.67   & 3.04 \\
            & DDPM (50)  & 0.78 & 0.78    &                      & 0.82   & 0.71   & 88.03   & 10.94 \\
            & MOTFM (10) & \textbf{0.82}  & \textbf{0.82}    &                      & \textbf{0.91}   & \textbf{0.84}   & \textbf{21.2}  & \textbf{3.65} \\
        \midrule
        \multirow{3}{*}{ResNet50} 
            & Real       & 0.88  & 0.88  & \multirow{3}{*}{UNet-Res50} & 0.92   & 0.87   & 15.02   & 2.84 \\
            & DDPM (50)  & 0.78    & 0.78    &                      & 0.73   & 0.59   & 133.18   & 25.66 \\
            & MOTFM (10) & \textbf{0.81}    & \textbf{0.81}    &                      & \textbf{0.91}   & \textbf{0.85}   & \textbf{20.89}   & \textbf{3.64} \\

        \bottomrule
    \end{tabular}
\end{table}

\begin{figure}[!h]
\centering
\includegraphics[width=1.0\textwidth]{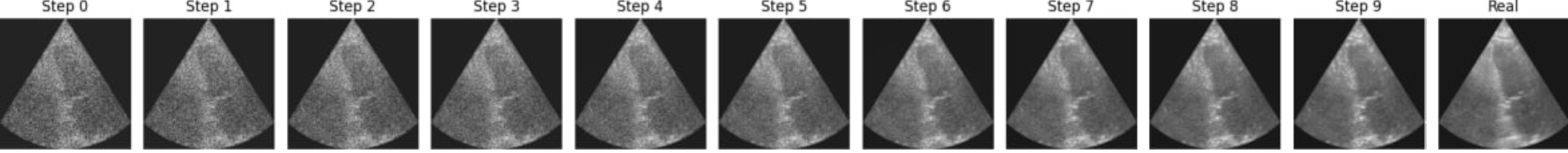} 
    \caption{Denoising Example: From a Noisy Image to a Denoised Image in 10 Steps}
    \label{denoising}
\end{figure}

\noindent\textbf{Generalizability to Other Medical Imaging Tasks.}
Our pipeline extends beyond image generation, demonstrating versatility in tasks like denoising. We apply it to speckle noise removal on the CAMUS dataset by adding speckle noise with different power to clean images and training the model to recover the original data. Instead of sampling from Gaussian noise, we initialize from noisy images, learning a transformation between noisy and clean domains (Fig.\ref{rf_diff}.b). While denoising is not our main focus, Fig.\ref{denoising} shows its promising performance across inference steps. Based on validation data, the denoised images achieve average metrics of PSNR: 32.76, SSIM: 0.8401, and SNR: 23.02, compared to the noisy images’ PSNR: 21.61, SSIM: 0.5802, and SNR: 11.87, highlighting its broader potential in medical imaging.

\section{Conclusion} 
This study introduces Optimal Transport Flow Matching framework for medical image synthesis with diverse conditioning strategies, adaptable across modalities and dimensions. Our approach surpasses diffusion-based baselines with fewer inference steps, achieving superior image quality and efficiency. Beyond synthesis, it can be extended to tasks such as image-to-image translation and denoising. Future work will focus on improving sample diversity with this pipeline.

\bibliographystyle{splncs04}
\bibliography{ref}

 \end{document}